\documentclass[]{article}
\usepackage{arxiv}
\usepackage[affil-it]{authblk}
\usepackage[english]{babel}

\usepackage{blindtext}
\usepackage[T1]{fontenc}
\usepackage{amsmath}
\usepackage{amssymb}
\usepackage{multirow}
\usepackage{tabu}
\usepackage{booktabs}
\usepackage[utf8]{inputenc} 
\usepackage{url}            
\usepackage{amsfonts}       
\usepackage{nicefrac}       
\usepackage{microtype}      
\usepackage{lipsum}
\usepackage{floatrow}
\usepackage{graphicx}
\usepackage{hyperref}
\usepackage{wrapfig}

\usepackage[label font=bf,labelformat=simple]{subfig}
\usepackage{caption}
\title{G-NeuroDAVIS: A Neural Network model for generalized embedding, data visualization and sample generation}

\author{Chayan Maitra}
\author{Rajat K. De*}

\affil{Machine Intelligence Unit, Indian Statistical Institute, 203 Barrackpore Trunk Road, Kolkata 700108, India.}

\affil{*Corresponding author: Rajat K. De (rajat@isical.ac.in)}

\begin{document}

\maketitle
\begin{abstract}
Visualizing high-dimensional datasets through a generalized embedding has been a challenge for a long time. Several methods have shown up for the same, but still, they have not been able to generate a generalized embedding, which not only can reveal the hidden patterns present in the data but also generate realistic high-dimensional samples from it. Motivated by this aspect, in this study, a novel generative model, called G-NeuroDAVIS, has been developed, which is capable of visualizing high-dimensional data through a generalized embedding, and thereby generating new samples. The model leverages advanced generative techniques to produce high-quality embedding that captures the underlying structure of the data more effectively than existing methods. G-NeuroDAVIS can be trained in both supervised and unsupervised settings. We rigorously evaluated our model through a series of experiments, demonstrating superior performance in classification tasks, which highlights the robustness of the learned representations. Furthermore, the conditional sample generation capability of the model has been described through qualitative assessments, revealing a marked improvement in generating realistic and diverse samples. G-NeuroDAVIS has outperformed the Variational Autoencoder (VAE) significantly in multiple key aspects, including embedding quality, classification performance, and sample generation capability. These results underscore the potential of our generative model to serve as a powerful tool in various applications requiring high-quality data generation and representation learning.
\end{abstract}

\keywords{Deep learning \and Unsupervised learning \and Generative model \and Data Visualization \and Conditional sample generation}


\section{Introduction}
\label{sec:intro}

Due to the rapid development of digital technologies, dealing with high-dimensional datasets is more relevant to get insights for data-driven decision-making \cite{DecisionMaking}. Many real-world problems such as data visualization, sample generation, image segmentation, and pattern recognition involve high-dimensional datasets for revealing entangled relationships \cite{HDData}. However, it becomes more challenging to visualize and capture information properly from those higher-dimensional datasets \cite{VisualIMP}. If the number of features increases, it leads to the problem of data sparsity, which might be an issue behind the poor performance of several machine learning models. In this context, managing high-dimensional datasets requires specialized techniques for dimensionality reduction and data visualization to explore meaningful hidden patterns \cite{methods}.

In recent years, many data visualization methods have come up with several ideas. Methods, like t-distributed Stochastic Neighbor Embedding (t-SNE) \cite{tSNE}, Uniform Manifold Approximation and Projection (UMAP) \cite{UMAP}, and IVIS \cite{IVIS} are well known for preserving the local structures, whereas Principal Component Analysis (PCA) \cite{PCA}, Singular Value Decomposition (SVD) \cite{SVD} and Multidimensional scaling (MDS) \cite{MDS} provide embedding that preserves global relationships over the local ones. An embedding that preserves local shapes over global ones is typically designed to maintain the relative distances and relationships between neighboring data points while avoiding broader patterns or variations in the overall dataset. In such embedding, closer points remain closer, which produces compact clusters and makes the embedding sparse. On the other hand, the embedding that preserves the global structures is not sparse but the cluster structure present in the data gets disrupted. In either case, the embedding does not have the generalization capability that shows a shortfall regarding sample generation from those embedding.

In order to address this problem of sample generation, several methods exist in recent literature. Variational Autoencoders (VAE) \cite{VAE}, Generative Adversarial Networks (GANs) \cite{GANs}, RealNVP (Real-valued Non-Volume Preserving) \cite{RealNVP}, and Glow (Generative Flow with Invertible 1x1 Convolutions) \cite{Glow} are capable of generating samples efficiently, but not always produce a latent embedding. VAEs are neural network-based models consisting of an encoder and a decoder. They can learn a latent representation from input data by minimizing the mean squared loss and the divergence between the learned latent distribution and a predefined prior distribution simultaneously. On the other hand, GANs, containing a generator and a discriminator network, are trained adversarially. The generator is able to map samples from a simple distribution (e.g., Gaussian noise) to the data space, while the discriminator learns to distinguish between real and generated samples. Through adversarial training, GANs produce high-quality samples that are similar to real data and can capture the underlying distribution. Flow-based models are another generative model that are able to learn a bijective mapping between a simple distribution (e.g., Gaussian) and the data distribution. They achieve it by modeling the data density through a sequence of invertible transformations. RealNVP and Glow are examples of flow-based models, which are also capable of generating high-quality images.

In this work, a feed-forward neural network model, called G-NeuroDAVIS, has been developed for data visualization through generalized embedding. In addition, the proposed network is also able to generate samples that are similar to real data. G-NeuroDAVIS takes an identity matrix as input and reconstructs the original data through a few hidden layers. During this reconstruction, it learns a close approximation of the actual data distribution using suitable parameters in the latent space. Once the the latent distribution is learned, by sampling from this distribution several samples in the high dimensional space are generated. This process of generation depends on the generalization capability of the latent space. Although this process can generate high-dimensional samples effectively, one does not have any control over these generations, i.e., all the generations are independent of each other. G-NeuroDAVIS has been trained in an unsupervised fashion, which stops it from conditional data generation. Motivated by this aspect we have modified the input using the label information and learned a conditional distribution at the latent space in a supervised fashion. This supervised model is efficient enough to generate samples based on a specific condition. Thus G-NeuroDAVIS is a model that can visualize data through a generalized embedding, produces samples in both conditional and unconditional scenarios, and can handle high-dimensional complex datasets. 

The performance of G-NeuroDAVIS is then compared against VAE, which has a similar capability. The embedding quality of VAE is compared against G-NeuroDAVIS first. The generalization capability and classification performance have been compared. Later the sample generation performance was evaluated against VAE. Finally, the performance of conditional data generation has been described. G-NeuroDAVIS has shown comparable performance in both the tasks of data visualization and sample generation.

The organization of the remaining part of this article is as follows. Section \ref{sec:methodology} discusses the detailed methodology of the proposed model. Section \ref{sec:Results} describes the experimental results, along with comparisons on several image datasets in terms of generalized embedding
, downstream analyses
, sample generation capability
, sub-sample based analyses
, and finally conditional sample generation performance
. Section \ref{sec:DisCon} provides a brief discussion of the strengths and weaknesses of the proposed model, and concludes this article.


\section{Methodology}
\label{sec:methodology}
In this study, we have developed a generative neural network model, called G-NeuroDAVIS, which is not only useful for visualizing high-dimensional datasets but also effective for generating new samples. This section describes the problem scenario and motivation behind the proposed model followed by its architecture, forward propagation, and learning.

\subsection{Problem scenario and solution approach}
\label{sec:problemsolution}
Here we narrate the problem scenario and solution approach.

\begin{flushleft}
    \textit{Problem scenario}
\end{flushleft}

The availability of high-dimensional datasets is growing rapidly due to the advancement of technology, but it becomes challenging to visualize and extract information effectively \cite{VisualIMP2}. If the number of features increases, the volume of the feature space grows exponentially, which leads to the problem of data sparsity and the distances among the samples become less informative. This is one of the major issues of the curse of dimensionality, which increases the computational complexity and model overfitting. In this context, managing high-dimensional datasets requires specialized techniques for dimensionality reduction and visualization to reveal meaningful patterns, relationships, and insights.   

Analogously, the other important aspect of machine learning is sample generation which involves creating synthetic data for effective training of the model, especially in scenarios where data is sparse, imbalanced, or missing \cite{sampleGeneration, sampleGeneration2}. For instance, in anomaly detection, generating realistic anomalous samples helps improve a model's ability to identify rare or outlier events. Similarly, in data imputation, generated samples fill in missing data points, enabling models to perform better in cases of incomplete data. In class-imbalance problems, generating synthetic samples from underrepresented classes helps prevent models from being biased toward the dominant class \cite{sampleGeneration, sampleGeneration2}. However, generating such samples is particularly challenging due to the complex and non-linear associations in high-dimensional feature spaces. High-dimensional data often exhibits intricate patterns, where features interact in non-obvious ways, making it difficult to replicate these relationships in synthetic samples. 

In this context, learning a generalized latent distribution is crucial in generative models because it allows for a structured, compact representation of complex data. This approach reduces the dimensionality of the data, facilitating efficient computation and meaningful sample generation. By mapping high-dimensional data into a lower-dimensional latent space, the model captures essential features and underlying structures, making it possible to generate new, diverse, and coherent samples by sampling from this latent distribution. Additionally, a generalized latent space supports smooth interpolation among data points, provides a regularizing effect to prevent overfitting, and offers interpretability for controlling and manipulating generated outputs. Overall, learning a generalized latent distribution enhances the model's ability to generate high-quality, varied, and realistic data while maintaining generalization and adaptability for various applications.

Moreover, an additional control on the sample generation process makes the task of generation even more useful. Sampling from the latent embedding can produce a random sample that is close enough to a real one. However, conditional data generation significantly enhances the sample generation process by incorporating additional contextual information or constraints into the model. This approach enables the generation of samples that meet specific requirements or attributes, leading to more targeted, relevant, and high-quality outputs. 

The main objective of the study is to develop a novel model that not only can visualize the data in lower dimensions but also generate samples from the original high-dimensional space simultaneously by learning the inherent distribution. In addition to that, the model can generate customized samples according to the targeted scenarios, which is one of the key goals of this study.

\begin{flushleft}
    \textit{Solution approach}
\end{flushleft}

Motivated by the previous model, viz., NeuroDAVIS \cite{neurodavis_neucom}, the proposed architecture of G-NeuroDAVIS can effectively utilized for both data visualization in lower dimensions along with the generation of samples in high-dimension. In the structure of the proposed model, assuming that the latent space follows a multivariate Gaussian, and the corresponding parameters, i.e., mean and variance have been learned accordingly. In this model, an identity matrix of order $n$ is fed as input, which controls the parameters of the Gaussian. Using these parameters, a sample has been drawn using the reparameterization trick \cite{VAE}. Once a sample is generated in the low dimensional space, it is projected into the original space through a few hidden layers to create a replica of the original data sample. In order to achieve two major targets, i.e., sample generation and generalization of the latent space, two loss functions have been considered here. The first one is a reconstruction loss or mean squared error loss which ensures that the projections or generations become realistic as well as close to the accurate value. On the other hand, KL-Divergence (KLD) loss has been calculated to ensure the generalization capability of the latent space. The KLD loss makes the latent distribution close to a standard normal distribution so that the latent space may not remain sparse. Using these two loss functions G-NeuroDAVIS has been trained on different datasets. Moreover, instead of feeding an identity matrix as input, a one-hot class matrix has been considered to control the sample generation process, so that one can customize the generated samples based on some pre-defined conditions. The detailed methodology for development of G-NeuroDAVIS is provided in the following section.

\subsection{G-NeuroDAVIS: Architecture, Forward Propagation and Learning}
\label{sec:method}
This subsection deals with the development of the architecture, forward propagation, and learning of the proposed G-NeuroDAVIS model. 

\begin{wrapfigure}{l}{0.5\textwidth}
\centering
\includegraphics[width=\textwidth]{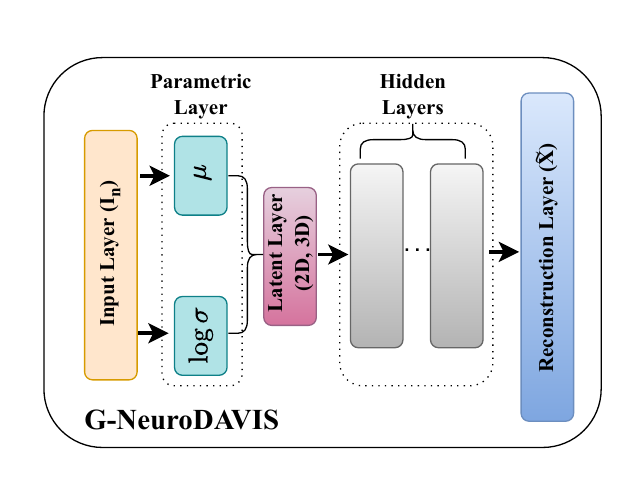}
\caption{\label{fig:Arch}The network architectures of G-NeuroDAVIS for visualization of high-dimensional datasets.}
\end{wrapfigure}

\begin{flushleft}
    \textit{Architecture}
\end{flushleft}
The architecture (Figure \ref{fig:Arch}) of the proposed G-NeuroDAVIS is a non-recurrent feed-forward neural network. The model can effectively extract a low-dimensional embedding which can efficiently capture the informative features for visualizing the high-dimensional datasets. G-NeuroDAVIS architecture contains five different types of layers, viz., an \textit{Input layer}, a \textit{Parametric layer}, a \textit{Latent layer}, one or more \textit{Hidden layer(s)} and a \textit{Reconstruction layer}.  

Let $\mathbf{X} = \{ \mathbf{x}_i: \mathbf{x}_i \in \mathbb{R}^d \}_{i=1}^{n}$ be a dataset comprising $n$ samples characterized by $d$ features each. The \textit{Input layer} of G-NeuroDAVIS consists of $n$ neurons/nodes. The \textit{Parametric layer} has two sub-parts, each having $k$ nodes, where $k$ is the number of dimensions of the projected space to be used for visualization. For the purpose of visualization, one should use $k=2$ or $3$. The \textit{Latent layer} contains $k$ nodes. The number of nodes in the \textit{Hidden layer(s)} depends on the dataset that needs to be visualized, and in this work, it has been set empirically. Moreover, the \textit{Reconstruction layer} has $d$ nodes, i.e., the same as the dimension of the original feature space. It has been noted that all the layers are densely connected, i.e., each node in any of these layers is connected to all the nodes of its adjacent layer(s). 

The \textit{Input layer} has been utilized to embed the original $n$ samples/observations independently into the lower dimensional space at the \textit{Latent layer} and in order to achieve the same a \textit{Parametric layer} has been considered. The \textit{Parametric layer} produces the mean and variance as parameters, which have been used further to define the distribution in the latent space. In the next stage, one or more \textit{Hidden layer(s)} have been used to project the lower dimensional embedding at the \textit{Latent layer} onto the \textit{Reconstruction layer}. Here, only two hidden layers have been taken into consideration. A \textit{Hidden layer} has been used to capture the essence of the non-linearity present in the dataset and carry the information to its next adjacent layer. In this context, multiple \textit{Hidden layers} can be used based on the computational complexity. However, the number of \textit{Hidden layers} should not be large enough as it tends to overfit the data. Consequently, the \textit{Reconstruction layer} tries to recreate the $n$ samples in the $d$ dimensional space using random low-dimensional embedding.

\begin{flushleft}
    \textit{Forward propagation}
\end{flushleft}
As stated above, the input of the proposed G-NeuroDAVIS is an identity matrix $\mathbf{I}_n$ for a dataset consisting of $n$ samples. An $i^{th}$ column vector $\mathbf{e}_{i}$ of $\mathbf{I}_n$ is fed to the input layer to create an approximate version of the $i^{th}$ sample $\mathbf{x}_i$ as $\mathbf{\tilde{x}}_{i}$, at the \textit{Reconstruction layer}. Let $\mathbf{a}^{(Inp)}_{i}$ and $\mathbf{h}^{(Inp)}_{i}$ represent the input and output of the \textit{Input layer}, respectively, for $i^{th}$ sample. Then 

\begin{equation}
	\begin{cases}
		\mathbf{h}^{(Inp)}_{i} = \mathbf{a}^{(Inp)}_{i} = \mathbf{e}_{i}, 
            \hspace{1cm} i=1,2,\cdots, n 
	\end{cases}
\end{equation}

Here, $\mathbf{e}_i$ controls the weight parameters for the lower dimensional projection of $i^{th}$ observation at the \textit{Latent layer}. Thus, the links connected to the $i^{th}$ neuron of the \textit{Input layer} can only activate the neurons in \textit{Parametric layer} on presentation of the $i^{th}$ sample.

As mentioned earlier, the \textit{Input layer} is densely connected with the \textit{Parametric layer}, which has two sub-parts. The first part calculates the mean ($\mu$) and the second part calculates the log-variance ($\log \sigma ^2$) of the distribution assumed at the latent space (specifically a multivariate Gaussian). Let $\mathbf{W}^{(\mu)}$ and $\mathbf{b}^{(\mu)}$ represent the weights and bias term respectively between the \textit{Input layer} and the first part of the \textit{Parametric layer}. Similarly, $\mathbf{W}^{(\sigma)}$ and $\mathbf{b}^{(\sigma)}$ represent the same between the \textit{Input layer} and the second part of the \textit{Parametric layer}. Assuming that for an $i^{th}$ sample, $\mathbf{a}^{(\mu)}_{i}$ and $\mathbf{a}^{(\sigma)}_{i}$, respectively, represent the input to the sub-parts of the \textit{parametric layer} before applying activation, whereas $\mathbf{h}^{(\mu)}_{i}$ and $\mathbf{h}^{(\sigma)}_{i}$, respectively, stand for the output of the same after activation. The linear activation has been used in both the sub-parts of the \textit{Parametric layer}. Hence, on presentation of an $i^{th}$ sample in the \textit{Parametric layer}, we have,

\begin{equation}
	\begin{cases}
		\mathbf{a}^{(\mu)}_{i} = \mathbf{W}^{(\mu)}\mathbf{e}_{i} + \mathbf{b}^{(\mu)},  \\ 
		\mathbf{h}^{(\mu)}_{i} = \mathbf{a}^{(\mu)}_{i}
	\end{cases}
    \begin{cases}
        \mathbf{a}^{(\sigma)}_{i} = \mathbf{W}^{(\sigma)}\mathbf{e}_{i} + \mathbf{b}^{(\sigma)}, \hspace{1cm} i=1,2,\cdots, n  \\ 
		\mathbf{h}^{(\sigma)}_{i} = \mathbf{a}^{(\sigma)}_{i}, \hspace{2.6cm} i=1,2,\cdots, n 
    \end{cases}
\end{equation}

As a result, the \textit{Parametric layer} takes the parameters of the multivariate Gaussian and produces a sample from it at the \textit{Latent layer}. However, the samples drawn directly from a distribution disrupt the training via gradient-based optimization. For this reason, the reparametrization trick has been used, which can separate the sampling process from the parameterization of the distribution \cite{VAE}. Instead of directly sampling from the distribution, sample selection from a standard Gaussian distribution was done at first and then transformed the same to match the desired distribution parameters. This transformation is differentiable and allows the gradients to flow through the sampling process. Let $\mathbf{a}^{(Lat)}$ and $\mathbf{h}^{(Lat)}$ correspond to the input to and output from the \textit{Latent layer}. Hence, for the \textit{Latent layer}, we have

\begin{equation}
	\begin{cases}
		\mathbf{a}^{(Lat)}_{i} = \mathbf{h}^{(\mu)}_{i} + exp(\frac{\mathbf{h}^{(\sigma)}_{i}}{2}) \odot \mathbf{\epsilon}_{i}, \hspace{1cm} i=1,2,\cdots, n \\ 
		\mathbf{h}^{(Lat)}_{i} = \mathbf{a}^{(Lat)}_{i} 
		\hspace{3.5cm} i=1,2,\cdots, n 
	\end{cases}
\end{equation}

Here, $\mathbf{\epsilon}_{i} \sim \mathcal{N}(\mathbf{0}, \mathbf{I})$, and $\odot$ represents element-wise multiplication. This sample $\mathbf{h}^{(Lat)}_{i} (\sim \mathcal{N}(\mathbf{h}^{(\mu)}_{i}, exp(\frac{\mathbf{h}^{(\sigma)}_{i}}{2})))$, produced at the latent layer, is used to reconstruct the $i^{th}$ sample $\mathbf{x}_i$. For this purpose, we have used \textit{Hidden layer(s)} and \textit{Reconstruction layer}. The sample generated from the \textit{Latent layer} is fed to the \textit{Hidden layer(s)} with a non-linear activation. Let $\mathbf{a}^{(Hid)}_{ij}$ and $\mathbf{h}^{(Hid)}_{ij}$, respectively, represent the input to and output from the $j^{th}$ \textit{Hidden layer} for the $i^{th}$ sample input. Thus, for $l$ \textit{Hidden layer(s)}, we have 

\begin{equation}
	\begin{cases}
	\mathbf{a}^{(Hid)}_{i1} = \mathbf{W}^{(Hid)}_{1}\mathbf{h}^{(Lat)}_{i}+\mathbf{b}^{(Hid)}_{1}, \hspace{1cm} i=1,2,\cdots, n\\ 
    \mathbf{h}^{(Hid)}_{i1} = ReLU(\mathbf{a}^{(Hid)}_{i1}), \hspace{2.25cm} i=1,2,\cdots, n\\
        \mathbf{a}^{(Hid)}_{ij} = \mathbf{W}^{(Hid)}_{j}\mathbf{h}^{(Hid)}_{i(j-1)} + \mathbf{b}^{(Hid)}_{j}, \hspace{0.85cm} i=1,2,\cdots, n; j=2,3,\cdots, l\\ 
		\mathbf{h}^{(Hid)}_{ij} = ReLU(\mathbf{a}^{(Hid)}_{ij}), 
		 \hspace{2.25cm} i=1,2,\cdots, n; j=2,3,\cdots, l
	\end{cases}
\end{equation}
where $ReLU(\mathbf{y}) = max(\mathbf{0},\mathbf{y})$; element wise \textit{max} (maximum) value has been considered. $\mathbf{W}^{(Hid)}_{j}$ and $\mathbf{b}^{(Hid)}_{j}$ correspond to the weight and bias connected to the $j^{th}$ \textit{Hidden layer(s)}. In this work, only two hidden layers have been considered.

A reconstruction of the original data is performed at the final layer, called \textit{Reconstruction layer}. For the \textit{Reconstruction layer}, we have

\begin{equation}
	\begin{cases}
		\mathbf{a}^{(Rec)}_{i} = \mathbf{W}^{(Rec)}\mathbf{h}^{(Hid)}_{il} + \mathbf{b}^{(Rec)}, \hspace{1cm} i=1,2, \cdots, n\\
		\mathbf{h}^{(Rec)}_{i} = \mathbf{a}^{(Rec)}_{i}, 
		\hspace{3.5cm} i=1,2, \cdots, n
	\end{cases}
\end{equation}
Here $\mathbf{W}^{(Rec)}$ and $\mathbf{b}^{(Rec)}$ correspond to the weight and bias term associated with the \textit{Reconstruction layer}.
Thus, G-NeuroDAVIS projects the latent embedding for a sample obtained at the \textit{Latent layer} into a $d$ dimensional space through the \textit{Hidden layer(s)}. 

\begin{flushleft}
    \textit{Learning}
\end{flushleft}
As mentioned earlier, G-NeuroDAVIS has been developed for dimensionality reduction as well as visualization of high-dimensional data. For each sample $\mathbf{x}_i$, ($i = 1,2, \cdots n$), G-NeuroDAVIS tries to find an optimal reconstruction $\mathbf{\tilde{x}}_i$ of the input data by minimizing the reconstruction error $\|\mathbf{x}_i-\mathbf{\tilde{x}}_i\|$. In addition, a KL Divergence (KLD) loss has been computed at the latent layer. KLD loss is defined as $p(x)\log(\frac{p(x)}{q(x)})$, where $p(x)$ and $q(x)$ are two distributions, and this loss gets minimized when $p(x)$ and $q(x)$ are identical. This regularizes the approximate posterior distribution to match a prior distribution (standard Gaussian $\mathcal{N}(\mathbf{0},\mathbf{I})$ in this case) over the latent space. It helps to follow a simple distribution, aiding generalization and facilitating interpolation in the learned latent space. Moreover, an $L2$ regularization may be used on the nodes' activities and edge weights to avoid over-fitting. The use of $L2$ regularization ensures the minimization of model complexity as well as the prevention of the weights from exploding. Hence, the complete objective function is

\begin{multline}
	\label{eqn:obj_func}
	\mathcal{L}_{G-NeuroDAVIS} = \frac{1}{n}\sum_{i=1}^{n}\| \mathbf{x}_{i} - \mathbf{\tilde{x}}_{i}\|_{2}^{2} + \beta  \sum_{i=1}^{n} KLD\Big(\mathcal{N}\big(\mathbf{h}^{(\mu)}_{i}, exp(\frac{\mathbf{h}^{(\sigma)}_{i}}{2})\big)||\mathcal{N}(\mathbf{0},\mathbf{I})\Big)+\\
    \gamma_1\sum_{j=1}^{l}\|\mathbf{W}^{(Hid)}_{j}\|_F^2+ \gamma_2\|\mathbf{W}^{(Rec)}\|_F^2+\gamma_3\sum_{j=1}^{l}\| \mathbf{h}^{(Hid)}_{ij} \|_2^2+ \gamma_4 \|\mathbf{h}^{(Rec)}_{i}\|_2^2,
\end{multline}

where $\beta$ is a balancing parameter, and $\gamma_1$, $\gamma_2$, $\gamma_3$ and $\gamma_4$ are regularization parameters; $\|.\|_2$ denotes the Euclidean norm of a vector, and $\|.\|_F$ indicates the Frobenius norm of a matrix. The balancing parameter $\beta$ helps to control the loss in a way such that both the objectives (i.e., reconstruction of samples and keeping the latent distribution close to a standard normal distribution) can be optimized simultaneously. Here, Adam optimizer has been used to minimize $\mathcal{L}_{G-NeuroDAVIS}$ in equation \eqref{eqn:obj_func} \cite{kingma2014adam}. Moreover, the number of epochs has been set empirically by observing saturation of $\mathcal{L}_{G-NeuroDAVIS}$.

At the initial stage of forward propagation, the weight values between the \textit{Input} and \textit{Parametric layer} have been fixed randomly. The learning of the proposed network is controlled by an $n^{th}$ order identity matrix $(\mathbf{I}_n)$ which has been fed as input to the \textit{Input layer}. The advantage of using the identity matrix confirms the necessary modifications in the weight values that are exclusively associated with the samples in the present batch. In the beginning, random parameters of the lower dimensional Gaussian for $n$ observations have been set at the \textit{Parametric layer}. Using these parameters in the \textit{Latent layer}, a latent embedding is produced and projected onto the \textit{Reconstruction layer} by the proposed model through a few \textit{Hidden layers}. A loss is then calculated using Equation \eqref{eqn:obj_func} and backpropagated up to the \textit{Input layer}. Correspondingly, the weight and bias values are updated for a better reconstruction of the samples of the present batch for the next forward pass. The process has been continued until the loss saturates. On completion of the training phase, the transformed feature set, extracted from the \textit{Latent layer}, is considered as the lower dimensional embedding of the data and is used for visualization for $k=2$ or $3$.

Once, the model is trained, it can generate new samples using the learned parameters of the latent distribution. For this purpose, a random point is sampled from the learned latent distribution and projected into the high-dimensional space using the learned parameters of \textit{Hidden layers} and \textit{Reconstruction layer}. The output from the \textit{Reconstruction layer} represents a generated sample that is very close to the original one but is not an exact copy. This randomness has been introduced during the sampling process.

\subsection{Conditional data generation in a supervised settings}
\label{sec:condition}
Conditional generation is important because it allows for the generation of data conditioned on specific attributes, contexts, or conditions. This capability adds flexibility and control to generative models, enabling them to produce outputs tailored to particular requirements or constraints. In this work, G-NeuroDAVIS has been trained in an unsupervised manner and thus falls short of conditional data generation. However, additional supervision about the sample types or class (condition) overcomes this situation.  

Earlier, in order to train G-NeuroDAVIS, an identity matrix of order $n$ is fed as an input. For reconstruction of an $i^{th}$ sample, only the weights connected to the $i^{th}$ input node get activated. This ensures that the task of reconstruction of all the samples has been performed independently. Therefore, an overall distribution of the samples has been learned, ignoring the class information. Thus when it comes to generate a new sample, the proposed model has no control over any prior conditions. In order to generate samples based on some pre-defined conditions, this input has now been changed to another sparse matrix which contains the conditional information about the samples. Let $n$ samples be present in the dataset and are distributed in $c$ distinct classes. Now a matrix $\mathcal{I}_{n \times c}$ has been constructed, which is also sparse in nature but contains the conditional information. Each row of this matrix has exactly one non-zero entry based on the class it belongs. Unlike the previous case, samples satisfying $c^{th}$ condition have been reconstructed using weights connected to the $c^{th}$ input node. In other words, now the samples are no longer independent, they are coming from the same distribution if they belong to the same class. The number of nodes in the \textit{Input layer} has been changed accordingly and the other layers kept the same as before. This model is again trained on the training data using equation \eqref{eqn:obj_func} as an objective function for a sufficient number of epochs. 

After successful training, G-NeuroDAVIS can be used for data generation based on a given condition. Suppose one needs to generate a sample that belongs to the $k^{th}$ class, then a vector of size $c$ is fed to the \textit{Input layer}, whose $k^{th}$ entry is $1$, and rest are zero. This input will control the weights accordingly and create a sample at the \textit{Latent layer} that belongs to the $k^{th}$ class.  



\section{Results}
\label{sec:Results}

This Section describes the details of the datasets used in this study along with the evaluation and result analyses.

\subsection{Data description}

In this work, four image datasets have been used to demonstrate the effectiveness of G-NeuroDAVIS model. These image datasets are \textit{MNIST} \cite{MNIST}, \textit{FMNIST} \cite{FMNIST}, \textit{OlivettiFaces} \cite{OlivettiFaces}, and \textit{COIL20} \cite{COIL20}. The details of the datasets have been described in Table \ref{tab:datasets}. \textit{MNIST} contains $60k$ grayscale images of handwritten digits of dimension $28\times 28$ each. \textit{FMNIST} contains $60k$ grayscale images of clothing of dimension $28\times 28$ each. In both cases, the number of samples is larger than the number of features, which may make the task easier. On the other hand, \textit{OlivettiFaces} contains only $400$ samples of $40$ different human faces of dimension $64\times 64$ each, and \textit{COIL20} comprises $1440$ samples of $20$ different objects of dimension $128 \times 128$ each. In the later two cases, the dimension is much larger than the number of samples, which makes the task of visualization even more complex. 

\begin{table}[ht]\footnotesize
\caption{Descriptions of the datasets used for evaluation of G-NeuroDAVIS}
\centering
\begin{tabular}{cccccc}
		\hline
		\textbf{Name} & \textbf{Description}                                                      & \textbf{\#Samples} & \textbf{\#Features} & \textbf{\#Classes} & \textbf{Source} \\ \hline
		$MNIST$         & grayscale images of handwritten digits                                    & 60,000             & 784                   & 10                  &    \cite{MNIST}          \\ 
		$FMNIST$        & grayscale images of clothing items                  & 60,000             & 784                   & 10                  &    \cite{FMNIST}         \\ 
		$OlivettiFaces$ & grayscale images of human faces                              & 400                & 4096                  & 40                  &    \cite{OlivettiFaces}          \\ 
		$Coil20$        & grayscale images of objects, captured from various angles    & 1440               & 16384                 & 20                  &    \cite{COIL20}          \\ \hline
\end{tabular}
\label{tab:datasets}
\end{table} 

Next, a few problem-specific preprocessing steps have been done for different datasets. At first, standard MinMax scaling has been done to normalize all the pixel values and then all the sample images have been flattened to a vector to fit in the model.  

\subsection{Result Analyses}

Initially, the embeddings produced by G-NeuroDAVIS, in an unsupervised manner, have been compared against those obtained by VAE
, and correspondingly, the generalization capability 
has been shown. In order to evaluate the efficiency of the embeddings, a classification task has been carried out for both embeddings
. On the other hand, new samples have been generated by the proposed model and VAE, and corresponding efficiency has been compared 
. Lastly, a sub-sample-based experiment has been conducted on both the datasets, viz., $MNIST$ and $FMNIST$ with $60k$ samples
. In addition, the performance of G-NeuroDAVIS has been observed in a supervised sense, in which it can generate samples based on a given condition 
.

\subsubsection{G-NeuroDAVIS in an unsupervised setting}
Here the performance of G-NeuroDAVIS in the context of unsupervised learning has been described. In other words, the results described in this section have been obtained by training G-NeuroDAVIS without having prior class information.

\begin{flushleft}
\textit{Visualization through generalized embedding}    
\end{flushleft}

Generalized embedding offers significant advantages in data analysis by simplifying complex, high-dimensional data into more manageable and informative lower-dimensional representations through dimension reduction.
The task of dimension reduction is crucial over the image datasets due to data sparsity, complex structure, and presence of noise. Data visualization is a more precise task of dimension reduction where the latent dimension is fixed to $2$ or $3$ so that the latent embedding can be visualized. G-NeuroDAVIS has been trained on the above-mentioned image datasets and a comparative analysis has been shown. 


\begin{figure}
\centering
\sidesubfloat[]{\includegraphics[width=0.96\columnwidth, height=5cm]{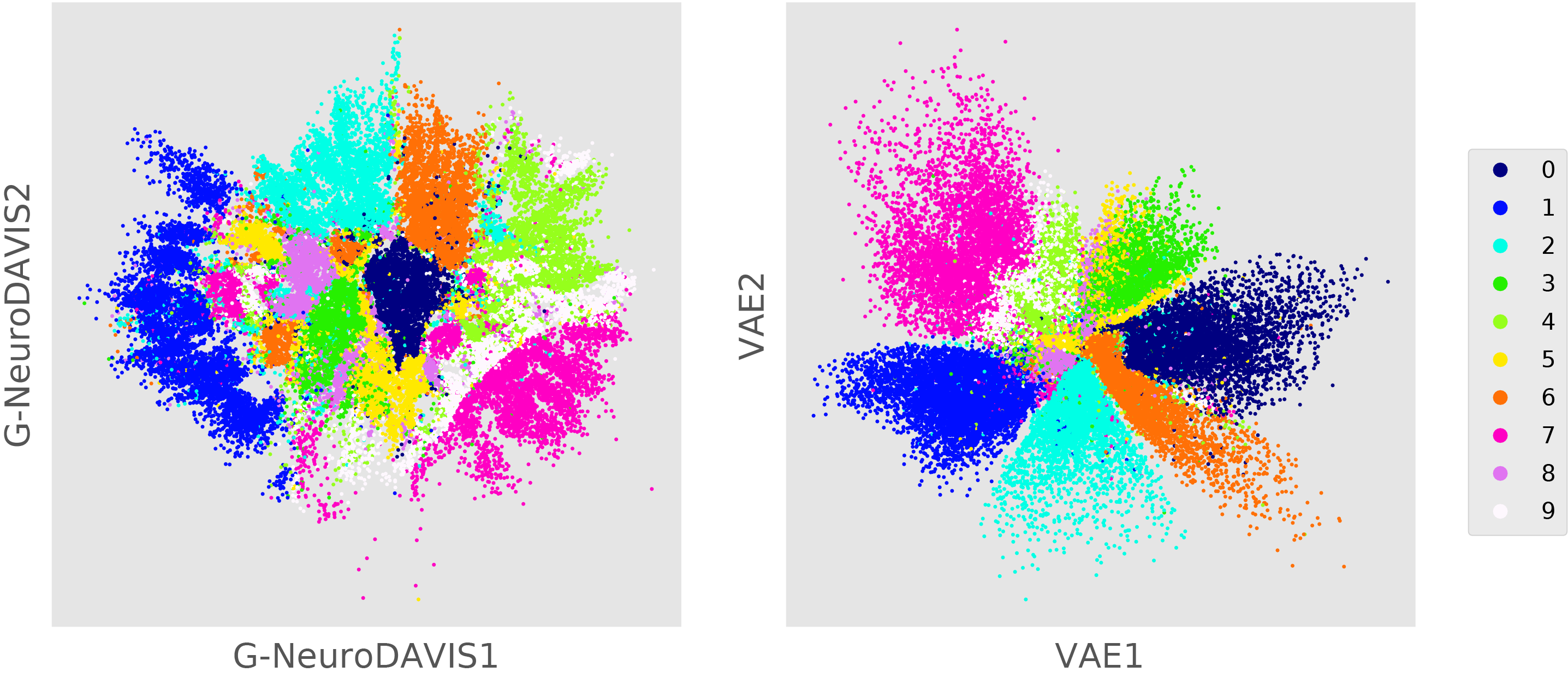}}\\[2ex]
\sidesubfloat[]{\includegraphics[width=0.96\columnwidth, height=5cm]{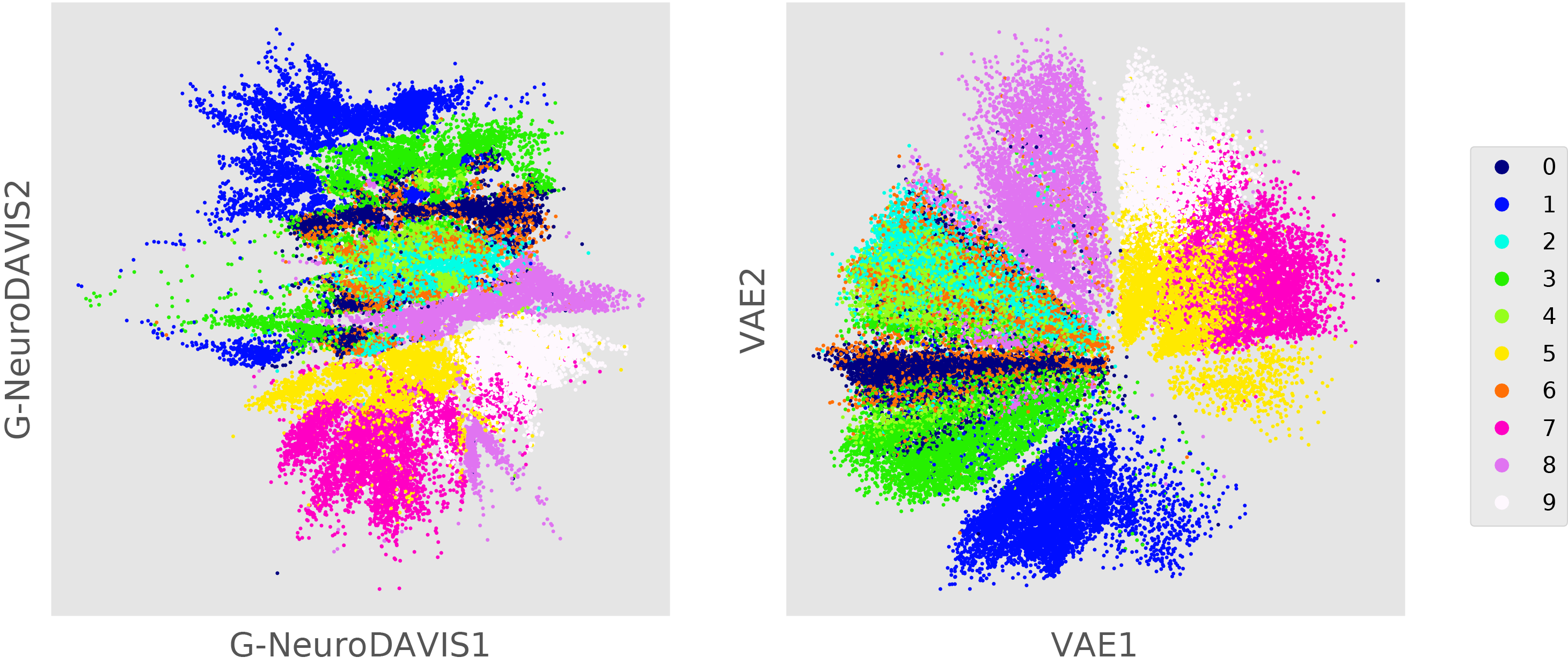}}
\caption{2-dimensional embeddings produced by G-NeuroDAVIS (left) and VAE (right) for (\textbf{A}) $MNIST$ and (\textbf{B}) $FMNIST$ datasets respectively, in the context of unsupervised learning (Different class samples are represented by different colors).}
\label{fig:embeddings_1}
\end{figure}

\begin{figure}
\centering
\sidesubfloat[]{\includegraphics[width=0.96\columnwidth, height=5cm]{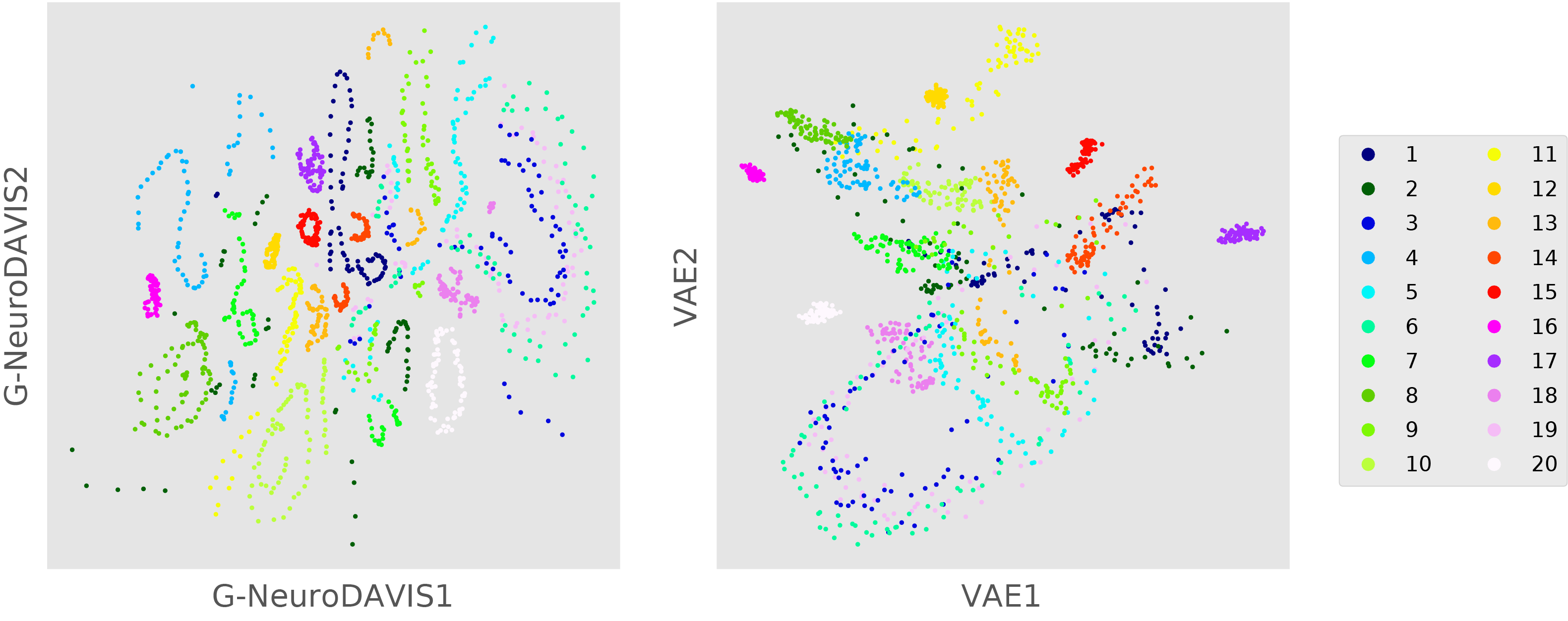}}\\[2ex]
\sidesubfloat[]{\includegraphics[width=0.96\columnwidth, height=5cm]{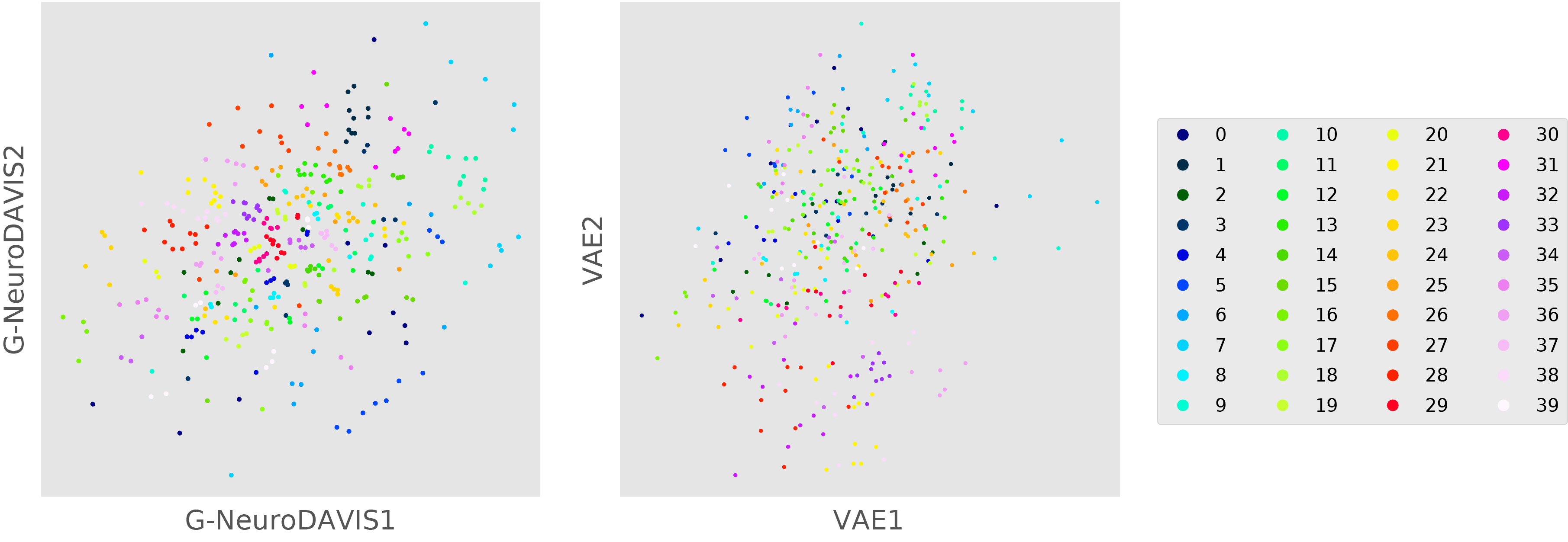}}
\caption{2-dimensional embeddings produced by G-NeuroDAVIS (left) and VAE (right) for (\textbf{A}) $Coil20$ and (\textbf{B}) $OlivettiFaces$ datasets respectively, in the context of unsupervised learning (Different class samples are represented by different colors).}
\label{fig:embeddings_2}
\end{figure}

In Figures \ref{fig:embeddings_1}A and \ref{fig:embeddings_1}B, $2-$dimensional embeddings, using G-NeuroDAVIS (left) and VAE (right) for $MNIST$ and $FMNIST$ datasets respectively, have been provided. The samples are colored with respect to different class labels. It has been revealed from the figures that in both the cases the classes are well separated. In the embeddings generated by G-NeuroDAVIS, different sub-clusters are observed, which is reasonable due to the intra-class variety present in both the datasets.

Likewise, $2-$dimensional embeddings obtained on $Coil20$ and $OlivettiFaces$ dataset using G-NeuroDAVIS (left) and VAE (right) have been shown in Figures \ref{fig:embeddings_2}A and \ref{fig:embeddings_2}B respectively. In $Coil20$ dataset, there are $72$  pictures of $20$ different objects each taken from different angles, which exhibit circular patterns in the original space. It has been revealed from Figure \ref{fig:embeddings_2}A that the embedding generated from G-NeuroDAVIS produces circular clusters, whereas, VAE generates compact clusters instead. It has been noted that in both the embeddings, similar points are closer. For the $OlivettiFaces$ dataset, the classes are well-separated and similar points remain closer in G-NeuroDAVIS-generated embedding. In contrast, VAE-produced embedding represents a mixture of different classes. This may happen due to the higher dimension of the dataset itself. 

Moreover, G-NeuroDAVIS-generated embeddings are generalized in nature. It reveals that the samples are densely positioned and each location in the embedding can be mapped to a realistic high-dimensional sample for each dataset. G-NeuroDAVIS is designed to enhance this generalization by learning compact, and informative representations that maintain the inherent semantic relationships of the original data. It outperforms VAE in terms of generalized embedding.

\begin{flushleft}
\textit{Embedding mapping}    
\end{flushleft}
In order to evaluate the effectiveness of the proposed model, G-NeuroDAVIS, an embedding mapping has been visualized and a comparison has been made against the same by utilizing VAE. In this context, a $30 \times 30$ grid space has been considered on the entire embeddings produced by both of them. Each point in the grid space is a possible lower dimensional point in the embeddings. These grid points have been projected to the high-dimensional space using the trained networks. These projections indicate the generation capability and diversity of the embeddings. These projections or reconstructions have then been fitted to the grid space to generate a pictorial representation of the learned mapping as well as capture the diversity.  

This experiment has been performed on all the above-mentioned datasets and the corresponding embedding mappings have been shown in Figure S1-S4 in the supplementary material. As stated earlier, small sub-clusters have been observed in the embedding generated by G-NeuroDAVIS in the $MNIST$ dataset, which has also been reflected in the embedding mapping (Figure S1 in supplementary material). 
From Figure S1 (left), it has been observed that the clusters, related to the number `$6$', occurred in two places; one at the top right corner and another at the bottom left corner. Though the numbers are the same in both the cases, the patterns are different. It has been observed that all different classes of data are present in the embedding mapping produced by G-NeuroDAVIS, whereas VAE has failed to do so. 

For $FMNIST$ dataset, there are no significant changes in the embedding mapping produced by the above-mentioned models (Figure S2 in supplementary material). It has been found from Figure S3 (supplementary material) that most of the clusters are well visible in G-NeuroDAVIS-generated embedding, whereas, VAE-generated embedding only reflects some major clusters for the mappings for $Coil20$ dataset.

Finally, for the $OlivettiFaces$ dataset, the embedding mapping produced by VAE is not at all capable of reconstructing the distinct faces. Instead, it learns only the common patterns present in the data (Figure S4 (right) in supplementary material). On the other hand, G-NeuroDAVIS is able to produce faces with unique features, such as glasses, beards, and different facial expressions. Moreover, each face-cluster is clearly visible in the compact embedding mapping (Figure S4 (left) in supplementary material) produced by G-NeuroDAVIS.

\begin{flushleft}
\textit{Downstream Analysis}    
\end{flushleft}


\begin{figure}
\centering
\sidesubfloat[]{\includegraphics[width=0.96\columnwidth]{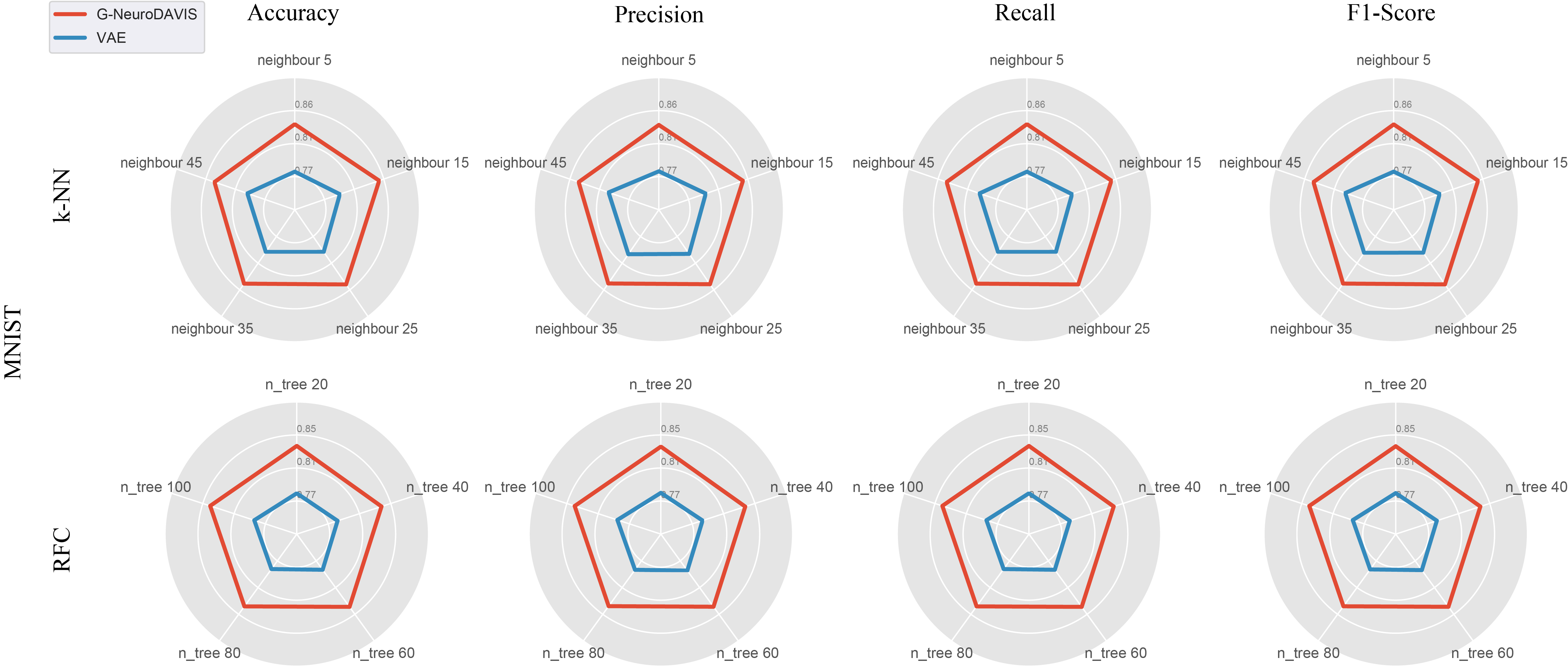}}\\[5ex]
\sidesubfloat[]{\includegraphics[width=0.96\columnwidth]{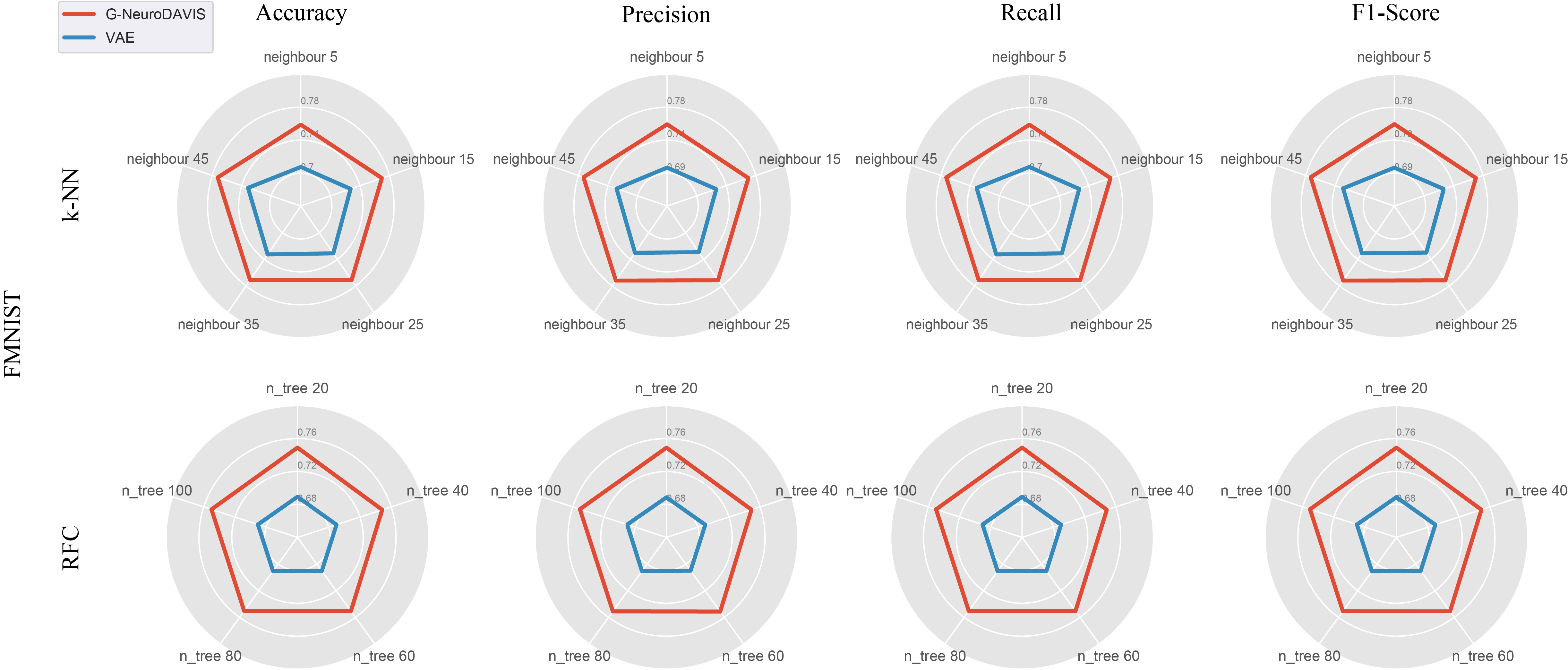}}\\[5ex]
\caption{Classification performance of G-NeuroDAVIS against VAE on (\textbf{A}) $MNIST$ and (\textbf{B}) $FMNIST$ dataset using k-NN and RF classifiers in terms of Accuracy, precision, Recall and F1-Score. Results have been obtained by varying parameter of k-NN ($neighbours$) and RF ($estimators$ / $trees$) respectively.}
\label{fig:classification_1}
\end{figure}

\begin{figure}
\centering
\sidesubfloat[]{\includegraphics[width=0.96\columnwidth]{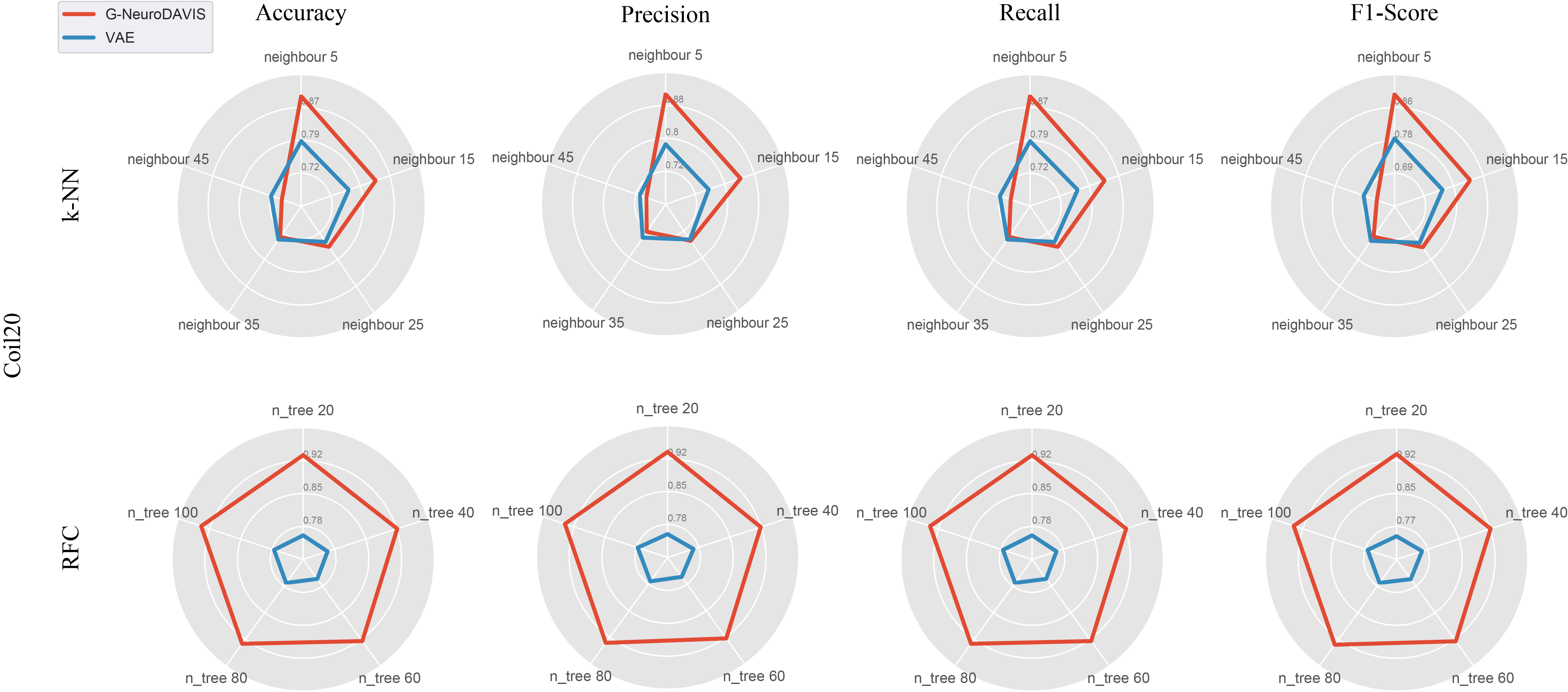}}\\[5ex]
\sidesubfloat[]{\includegraphics[width=0.96\columnwidth]{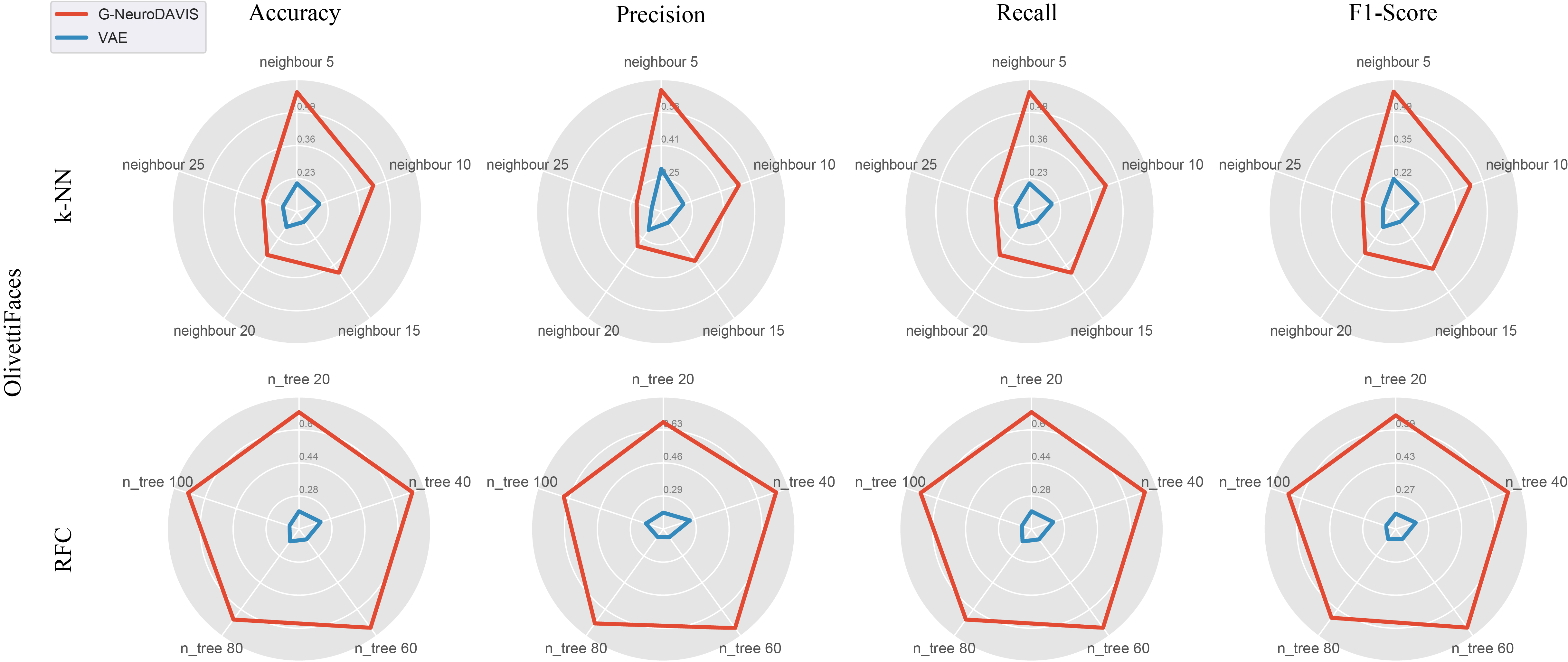}}\\[5ex]
\caption{Classification performance of G-NeuroDAVIS against VAE on (\textbf{A}) $Coil20$ and (\textbf{B}) $OlivettiFaces$ dataset using k-NN and RF classifiers in terms of Accuracy, precision, Recall and F1-Score. Results have been obtained by varying parameter of k-NN ($neighbours$) and RF ($estimators$ / $trees$) respectively.}
\label{fig:classification_2}
\end{figure}

\begin{figure}
\centering
\includegraphics[width=0.96\textwidth]{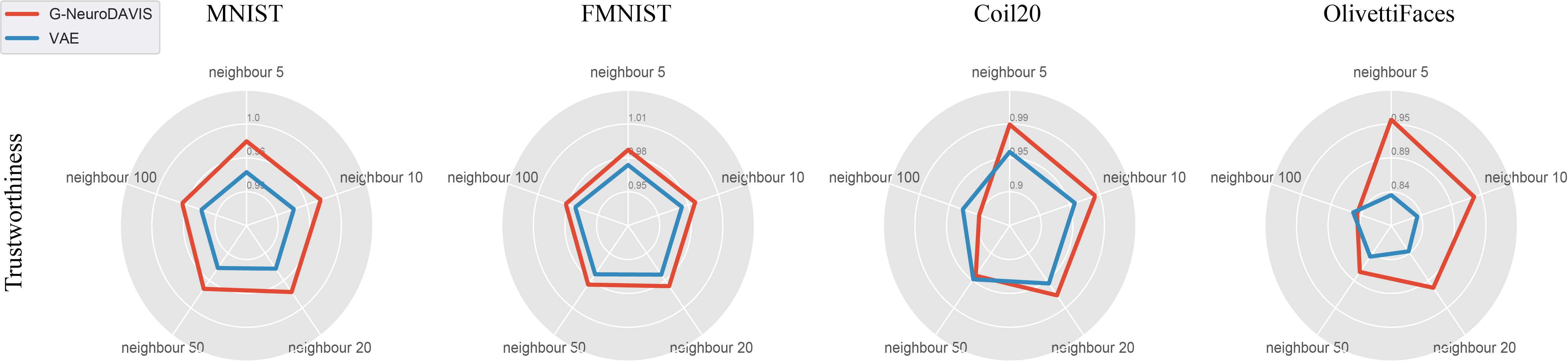}\\[2ex]%
\caption{Trustworthiness scores obtained on G-NeuroDAVIS-generated embedding of $MNIST$, $FMNIST$, $Coil20$, and $OlivettiFaces$ datasets and compared against that obtained by using VAE. Parameter $n\_neighbour$ has been varied for robust results.}
\label{fig:Trust}
\end{figure}

After successful training of both G-NeuroDAVIS and VAE, embeddings have been obtained from the latent layer of G-NeruoDAVIS and the bottleneck layer of the VAE respectively. These embeddings have again been compared with respect to several downstream analyses. Initially, the classification performance of both the models has been compared over all the datasets. In this study, k-NN and Random Forest (RF) classifiers have been utilized. For this comparison, all these datasets have been split into a $80:20$ ratio in which the larger part is used to train the classifiers, and the remaining part is then used to test the classifiers. Accuracy, Precision, Recall, and F1-scores have been reported to quantify the classification performance. For both $MNIST$ and $FMNIST$ datasets, G-NeuroDAVIS has outperformed VAE in terms of the above-mentioned metrics (Figure \ref{fig:classification_1}).

For $Coil20$ dataset, RF shows superior performance of G-NeuroDAVIS over VAE, but k-NN shows comparable performance for a higher value of the parameter $k$ (Figure \ref{fig:classification_2}A). On the contrary, both k-NN and RF show superior performance of G-NeuroDAVIS-generated embedding over VAE-generated embedding for $OlivettiFaces$ dataset in terms of Accuracy, Precision, Recall, and F1-score (Figure \ref{fig:classification_2}B). This experiment shows that G-NeuroDAVIS generated latent space has been able to preserve the class information though it has been trained in an unsupervised setting. The classifiers have been trained by varying their parameters to check their robustness.  

Later, we have evaluated the embedding quality using another well known metric, called Trustworthiness. It takes both the embedding and the original high-dimensional dataset as input and calculates how close the embedding is in terms of local distances. Locality has been measured by a parameter, called $n\_neighbour$. In this study, we have used five different values for $n\_neighbour$ and observed superior performance of G-NeuroDAVIS over VAE for $MNIST$ and $FMNIST$ datasets. For $Coil20$, the results obtained are quite comparable, and for $OlivettiFaces$, a higher score has been observed for G-NeuroDAVIS for lower values of $n\_neighbour$, as the value of $n\_neighbour$ increases similar result has been found \ref{fig:Trust}.

\begin{flushleft}
\textit{Sample Generation}    
\end{flushleft}


\begin{figure}
\centering
\includegraphics[width=0.96\textwidth]{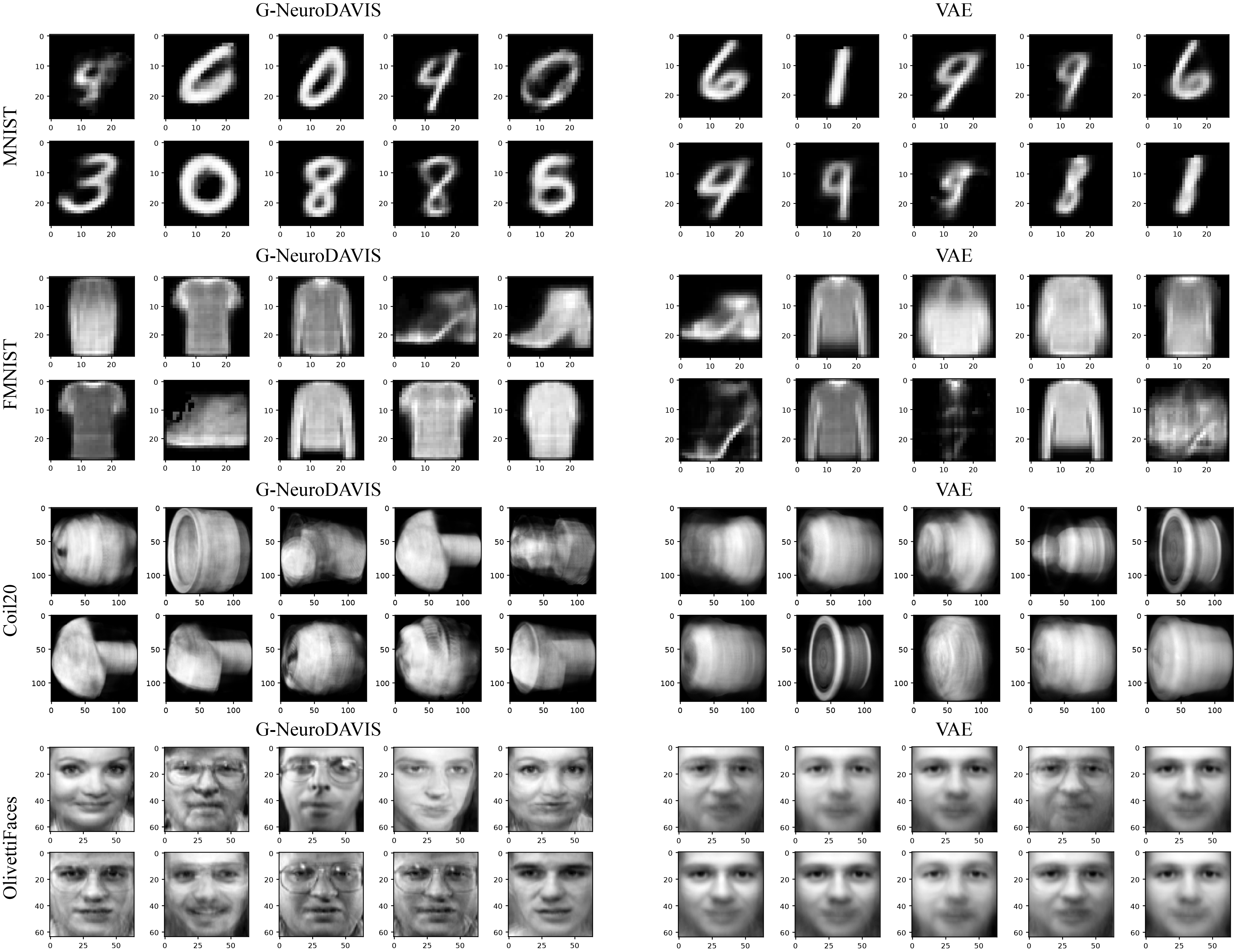}%
\caption{Samples generated using G-NeuroDAVIS (left) and VAE (right) in an unsupervised fashion on $MNIST$, $FMNIST$, $Coil20$, and $OlivettiFaces$ dataset.}
\label{fig:generation}
\end{figure}

This section presents a comparison between G-NeuroDAVIS and VAE in terms of sample generation capability. A latent embedding is created after the training of G-NeuroDAVIS, and this latent embedding can project sample points into a high dimensional space of the original dataset. For the same, a random point is first selected from the embedding's range space, and it is then projected to a high dimension by utilizing the hidden layers and reconstruction layer to create a sample from the high dimensional space. For validation purposes, $10$ such samples from the corresponding embeddings for each of the four datasets have been taken into consideration. 

As seen in Figure \ref{fig:generation}, G-NeuroDAVIS-generated samples for $MNIST$ dataset are realistic and readable; for $FMNIST$ dataset, the generated samples by G-NeuroDAVIS show distinct clothing items; for $Coil20$ dataset, various objects have been found clear in the G-NeuroDAVIS-generated samples; and for $OlivettiFaces$ dataset, G-NeuroDAVIS has been found to generate distinct faces. On the other hand, the generation from VAE model is not at all comparable for datasets with very high dimensions, like $Coil20$ and $OlivettiFaces$. For $OlivettiFaces$, all generated samples of facial expression contain only common features, but could not capture the unique features present in the data; and for $Coil20$, the samples are blurred in nature. However, for $MNIST$ and $FMNIST$ datasets comparable samples has been observed.

\begin{flushleft}
\textit{Sub-sample based analyses}    
\end{flushleft}
Another sub-sample-based experiment has been carried out to validate the robustness of the proposed G-NeuroDAVIS model against VAE, by altering the sample amount. For the same, different sizes of random subsamples ($10\%$, $20\%$, $\cdots$, $90\%$, entire data) of $MNIST$ and $FMNIST$ datasets with a sample size of $60k$ have been taken into consideration. It has been found that all the sub-samples are from a similar distribution (Figures S5 and S6 in supplementary material). 

Apart from visualization, a classification performance has been obtained on the embedding using k-NN and RF by altering their parameters. Figures S7 and S8 (in the supplementary material) demonstrate that, for all embeddings of varying sizes, G-NeuroDAVIS-produced embeddings have outperformed that by VAE in terms of accuracy, prediction, recall, and F1-Score.
 

\begin{figure}
\centering
\includegraphics[width=0.96\textwidth]{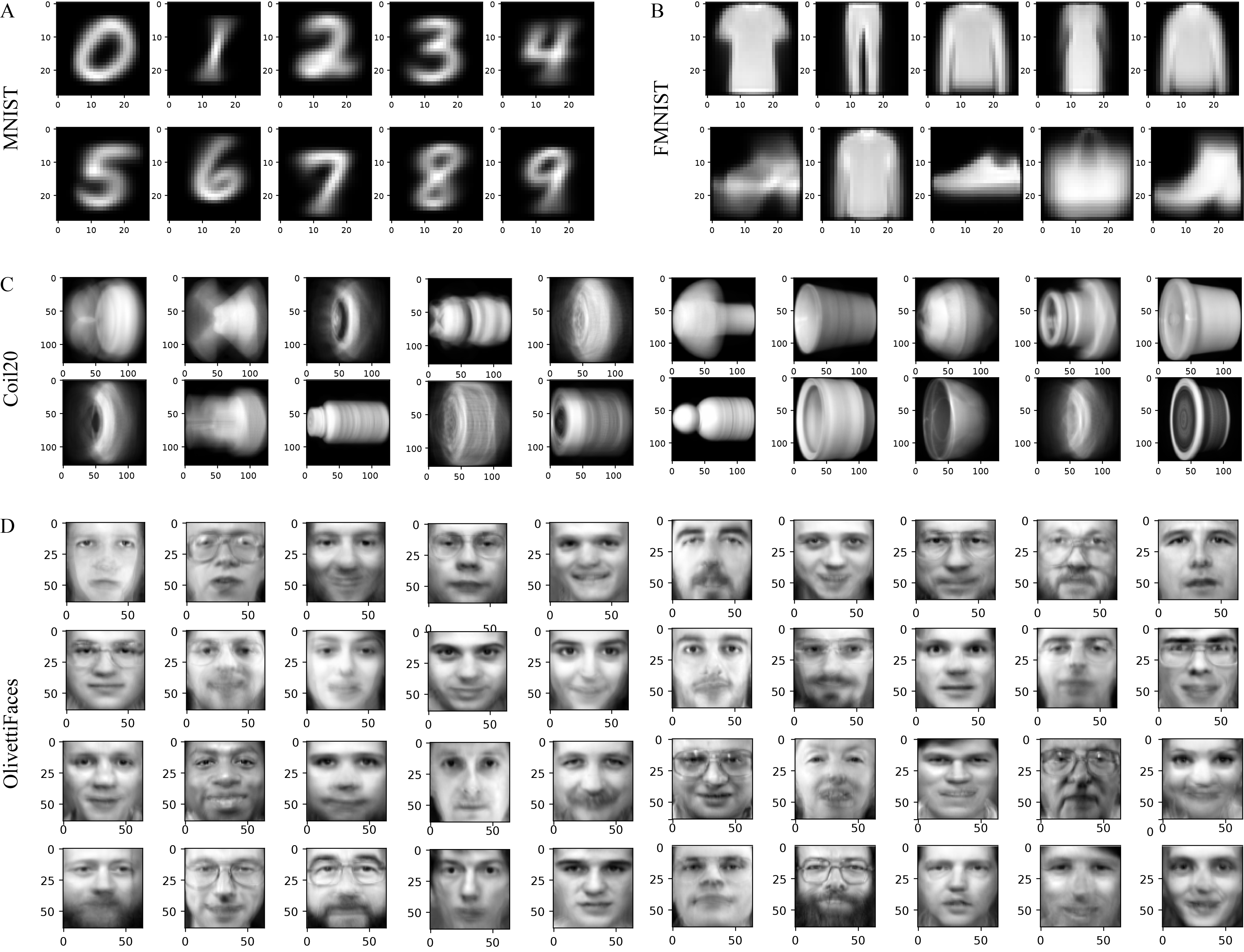}%
\caption{Samples generated using G-NeuroDAVIS in an supervised fashion on (\textbf{A}) $MNIST$, (\textbf{B}) $FMNIST$, (\textbf{C}) $Coil20$, and (\textbf{D}) $OlivettiFaces$ dataset.}
\label{fig:generation_sup}
\end{figure}

\subsubsection{G-NeuroDAVIS in a supervised setting}
\label{sec:Supervised}

Unlike other generative models, G-NeuroDAVIS can be trained in a supervised setting also and has been used for the above-mentioned datasets. G-NeuroDAVIS learns a distribution specifically for each class, based on the label information provided from the data itself. This learning technique enables conditional data generation. Initially, an embedding has been produced after the model training (Figure S9 in supplementary material) and samples have been generated accordingly. In the context of unsupervised learning, no information is given about the classes, and as a result, some overlapping of classes has been observed in Figures \ref{fig:embeddings_1} and \ref{fig:embeddings_2}. However, after having the class information, G-NeuroDAVIS has produced well-separated distinct classes across all the datasets (Figure S9 in supplementary material).

In addition, its conditional data generation capability has been examined, and accordingly, the samples have been generated from all the classes present in the dataset. For each class one sample has been generated using the trained model and all the conditional samples have been depicted in Figure \ref{fig:generation_sup}. 

Conditional data generation validation involves ensuring that the generated data meets specific conditions or constraints. It has been observed from Figure \ref{fig:generation_sup}, clear readable images for $MNIST$; all the clothing items, including bag and shoes for $FMNIST$; for $Coil20$ dataset, all $20$ distinct items; and for $OlivettiFace$ dataset, all $40$ different faces have been generated. Each of these generations matches the specific condition very well. G-NeuroDAVIS has been able to generate those class-specific samples successfully, not only for datasets having less number of classes but also for datasets having more than $20$ different classes.

This supervised learning adds a control to the generation capability of G-NeuroDAVIS and makes it even more useful than the state-of-the-art methods. 


\section{Discussion and Conclusion}
\label{sec:DisCon}
In this study, the authors have developed a feed-forward neural network model, called G-NeuroDAVIS, for data visualization through generalized embedding, and sample generation. G-NeuroDAVIS has been able to produce meaningful embedding by extracting features from high-dimensional datasets. The embedding, produced by it, can generate new valid samples as well. G-NeuroDAVIS can be trained in both supervised and unsupervised setups. Training in a supervised setting allows it to generate samples based on some pre-defined condition like classes, and in this way, one can have control over the generations. 

The performance of G-NeuroDAVIS has been compared against VAE, which has a similar capability, over four publicly available image datasets. Unlike VAE, G-NeuroDAVIS can generate samples based on conditions. It has been observed that though the embedding produced by G-NeuroDAVIS is similar to VAE in terms of visualization, the classification performance reflects superiority over VAE. The sub-sample-based experiment also shows the robustness of G-NeuroDAVIS over VAE. G-NeuroDAVIS has also shown superior generation capability over VAE.

G-NeuroDAVIS though has performed well on various datasets but has a few weaknesses. First, the input it takes to visualize the data is of the order of the square of the sample size, which makes a huge space complexity. Second, being a neural network model, it has several hyper-parameters that need to be set wisely. Sometimes a little modification may affect the results. Finally, G-NeuroDAVIS assumes the embedding to follow a Gaussian distribution, which may not be the case always. 

	
\section*{Authors' contributions statement}
Conceptualization of methodology and framework, Data curation, Data analysis, Implementation, Initial draft preparation, Investigation, Validation: CM. Reviewing, Editing, Overall Supervision: RKD.

\section*{Data and Code availability}
Codes to reproduce the results can be found at \url{https://github.com/shallowlearner93/G-NeuroDAVIS}.

\bibliography{mybib.bib}

\end{document}